\documentclass[10pt,twocolumn,letterpaper]{article}

\usepackage{cvpr}              % To produce the CAMERA-READY version
\usepackage{graphicx}
\usepackage{amsmath}
\usepackage{amssymb}
\usepackage{booktabs}
\usepackage{multirow}
\usepackage{bbm}
\usepackage{amsfonts}
\usepackage{xcolor}
\usepackage[title]{appendix}

\usepackage[pagebackref,breaklinks,colorlinks]{hyperref}

\usepackage[capitalize]{cleveref}
\crefname{section}{Sec.}{Secs.}
\Crefname{section}{Section}{Sections}
\Crefname{table}{Table}{Tables}
\crefname{table}{Tab.}{Tabs.}

\def\confName{CVPR}
\def\confYear{2024}

\begin{document}

\title{Make-A-Storyboard: A General Framework for Storyboard \\
with Disentangled and Merged Control}

\author{Sitong Su\footnotemark[1]\hspace{-0.2cm}\and
Litao Guo\footnotemark[1]\hspace{-0.2cm}\and
Lianli Gao\hspace{-0.2cm}\and
Heng Tao Shen\hspace{-0.2cm}\and
Jingkuan Song \and 
University of Electronic Science and Technology of China (UESTC)
\\
{\tt\small {sitongsu9796@gmail.com}}
}

\twocolumn[{%
\renewcommand\twocolumn[1][]{#1}%
\maketitle
\begin{center}
    \centering
    \includegraphics[width=0.95\linewidth]{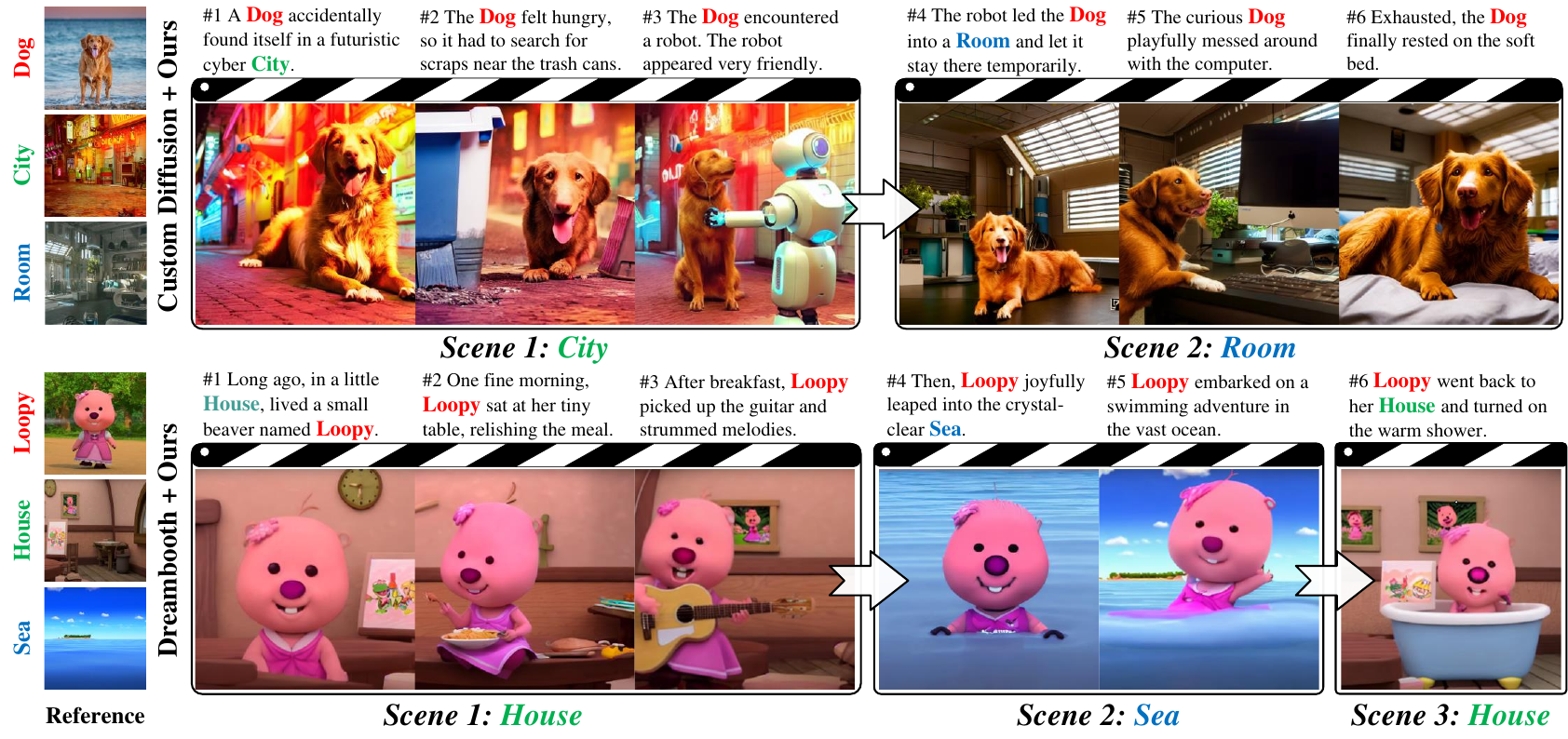}
    \captionof{figure}{\textbf{Demonstration of Storyboard.} We present a new representation form called Storyboard for Story Visualization, which unfolds contents of a story scene by scene.
    As shown, ours could be seamlessly integrated into Image Customization methods, enabling story visualization with contextual consistent characters and scenes.}
    \label{frontpage}
\end{center}%
}]

\renewcommand{\thefootnote}{\fnsymbol{footnote}} 
\footnotetext[1]{These authors contributed equally to this work.}

%%%%%%%%% ABSTRACT
\begin{abstract}
Story Visualization aims to generate images aligned with story prompts, reflecting the coherence of storybooks through visual consistency among characters and scenes.
Whereas current approaches exclusively concentrate on characters and neglect the visual consistency among contextually correlated scenes, resulting in independent character images without inter-image coherence.
To tackle this issue, we propose a new presentation form for Story Visualization called \textbf{Storyboard}, inspired by film-making, as illustrated in Fig.~\ref{frontpage}.
Specifically, a Storyboard unfolds a story into visual representations \textbf{scene by scene}~\cite{storyboard}.
Within each scene in Storyboard, characters engage in activities at the same location, necessitating both visually consistent scenes and characters.
For Storyboard, we design a general framework coined as Make-A-Storyboard that applies disentangled control over the consistency of contextual correlated characters and scenes and then merge them to form harmonized images.
Extensive experiments demonstrate \textbf{1) Effectiveness.} the effectiveness of the method in story alignment, character consistency, and scene correlation; \textbf{2) Generalization.} Our method could be seamlessly integrated into mainstream Image Customization methods, empowering them with the capability of story visualization.

\end{abstract}

%%%%%%%%% BODY TEXT
\section{Introduction}
\label{sec:intro}
\textit{A storyboard is a visual representation of how a story will play out, scene by scene.}---Luke Leighfield

Story Visualization synthesizes a series of images according to story prompts for better storybook illustration, which unleashes the imagination of readers and favoring interactive education. To reflect contextual coherence of story prompts, visual appearance of same characters and contextual correlated scenes shall remain consistent. To raise the story of Frozen as an example, the prompt ``Elsa lives in a Crystal Castle" implies that the visual appearance of the ``Elsa" and ``Crystal Castle" both requires consistency.

Considering the applicability of Story Visualization, several works have been dedicated to this field recent years.
Early works are trained on closed-domain datasets, which severely restrict the generalization~\cite{li2019storygan, maharana2021improving, maharana2022storydall, pan2022synthesizing, rahman2023make}.
Inspired by open-domain Text-to-Image (T2I) models~\cite{ho2020denoising,nichol2022glide,ldm}, subsequent works~\cite{gong2023talecrafter, liu2023intelligent} extend pretrained T2I models into massive story corpora and therefore achieve generalized visualization. To accommodate resource-friendly story visualization, ~\cite{jeong2023zero} introduces zero-shot Story Visualization based on coherent story descriptions. 

Despite the remarkable progress in the aforementioned works, they only attend to the visual consistency of characters while neglecting the contextual correlation of scenes. Consequently, scenes of the synthesized images are unrelated, leading to a series of independent character images without logical coherence among images.

To address this issue, we draw inspiration from the process of film-making and present a new representation form for Story Visualization called \textbf{Storyboard}, which plays out story contents \textbf{scene by scene}, as illustrated by Fig.~\ref{frontpage}.
In details, in each scene of Storyboard, characters involve in activities without the change of places, thereby demanding for consistent scene representation.
By employing Storyboard, the coherence and expressiveness of the Story Visualization is significantly enhanced because visual consistency within each scene is ensured and story progression is explicitly demonstrated to readers through transitions among scenes.  

To achieve Storyboard, we propose a general framework named \textbf{Make-A-Storyboard} which applies disentangled control over the consistency of scenes and characters, and then merge them through spatial masks to compose harmonized and visually consistent images.
The pipeline of our method is demonstrated in Fig.~\ref{pipeline}. \textbf{Firstly}, in Contextual Prompt Processing, LLMs are adopted for sorting out contextual relations in story prompts and process prompts to acquire specific character concepts and scene concepts.
\textbf{Secondly}, disentangled schedules by Image Customization methods~\cite{ruiz2023dreambooth, kumari2023multi} are employed to respectively learn the visual concepts of characters and scenes.
\textbf{Thirdly}, to merge characters and scenes in a reasonable way, we make insightful observations on fusing features from two disentangled T2I schedules by experiments on StableDiffusion~\cite{ldm} as shown in Fig.~\ref{observation}.
To raise the first row of Fig.~\ref{observation} as an example, ``squirrel" and ``pine forest" are merged by the squirrel mask at different timesteps of the denoising process. We observe that from pixel side to noise side the merged image gradually transits from a total cut-and-paste effect to a totally text-conditioned image unrelated to the original squirrel image.
\textbf{Lastly,} inspired by the observation, we design the Balance-Aware Merge Strategy that merges the features from the character and scene branches in the middle of the denoising process to balance harmonization and visual consistency.

The major contributions can be summarized as follows:

1) We present a new representation form for Story Visualization coined as Storyboard, which unfolds a story scene by scene, thus significantly enhancing coherence and expressiveness of Story Visualization.

2) For Storyboard, we design a general framework called Make-A-Storyboard which employs disentangled control on contextual correlated characters and scenes, and merges them to form harmonized images.

3) Extensive experiments demonstrate that our method can be seamlessly integrated into mainstream Image Customization methods, achieving both character consistency and scene correlation.

\section{Related Works}

\subsection{Story Visualization}
The concept of story visualization is initially introduced by StoryGAN~\cite{li2019storygan}. Many subsequent works also follow the adversarial training scheme~\cite{maharana2021improving,maharana2021integrating,li2022word}. Additionally, there are endeavors to enhance visual quality and semantic alignment through transformer-based methods~\cite{chen2022character, maharana2022storydall}. With the impressive generative capabilities demonstrated by diffusion models, works~\cite{pan2022synthesizing, feng2023improved} like Make-A-Story~\cite{rahman2023make} introduce a historically-aware autoregressive latent diffusion model. Nevertheless, the generalization ability of the above works remains constrained by fixed datasets such as Pororo~\cite{li2019storygan} and Flintstones~\cite{gupta2018imagine}.

Addressing this limitation, recent developments in open-domain story visualization emerge. ~\cite{jeong2023zero} focuses on the issue of identity consistency for novel characters; however, its applicability is limited to human faces and struggles to extend to other objects, with suboptimal image consistency. Talecrafter~\cite{gong2023talecrafter} introduces the C-T2I module, emphasizing character identity preservation and image layout. Coinciding with Talecrafter's timeline, StoryGen~\cite{liu2023intelligent} introduces open-domain visual storytelling and curates a large-scale storybook dataset. However, both Talecrafter~\cite{gong2023talecrafter} and StoryGen~\cite{liu2023intelligent} necessitate training their model components on extensive and intricate datasets.
However, all the current open-domain story visualization approaches solely focus on character identity consistency while overlooking the significance of scene consistency.

\subsection{Image Customization and Composition}

Remarkable success is attained by Text-to-Image(T2I) generation models relying on GANs~\cite{brock2018large,goodfellow2020generative,karras2021alias,karras2019style} and VAEs~\cite{van2017neural,kingmaauto}. Lately, diffusion-based models~\cite{dhariwal2021diffusion,ho2020denoising,sohl2015deep,nichol2022glide,ramesh2022hierarchical,saharia2022photorealistic,yu2022scaling,gu2022vector, ldm}  demonstrate amazing performance and become the mainstream approaches in T2I models.

Despite the general capability of these T2I models, users usually need to generate images based on specific requirements. On the one hand, Image Composition methods~\cite{avrahami2022blended,avrahami2023blended,chefer2023attend,feng2022training,liu2022compositional,lu2023tf,jin2023image} show significant efficacy in editing based on given images. However, it falls short in altering character actions and poses according to prompts. This renders Image Composition unsuitable for story visualization.
On the other hand, Image Customization methods~\cite{gal2022image,ruiz2023dreambooth,wei2023elite,shi2023instantbooth,hao2023vico,kumari2023multi,ma2023subject,xiao2023fastcomposer,jin2023image} fine-tune several similar images given as input, thereby achieving better alignment with the textual prompts. However, many current Image Customization approaches struggle to simultaneously fine-tune multiple concepts.
Although some multiple-concepts Image Customization methods~\cite{kumari2023multi,ma2023subject,xiao2023fastcomposer,jin2023image} try to handle multiple concepts, their performance in simultaneously managing scenes and characters is not satisfying. In contrast, our proposed framework could be seamlessly integrated into mainstream Image Customization method achieving scene and character alignment.

\section{Preliminaries}

\textbf{Stable Diffusion.}
% We select Stable Diffusion (SD)~\cite{ldm} as our foundational network architecture.
During the training phase, SD~\cite{ldm} initially compresses the input image $X$ in pixel space through a VAE encoder $E(\cdot)$ into the latent space as $x_0$. During the total $T$ steps of the forward process, noise gradually is injected into $x$. Formally, the forward process $q(x_t|x_{t-1})$ is:
\begin{equation}
q(x_t|x_{t-1}) = \mathcal{N}(x_t; \sqrt{1-\beta_t}x_{t-1}, \beta_tI) , t = 1,...,T
\end{equation}
where  ${\{\beta_i\}}^T_{i=1}$ is the variance schedule hyperparameters.
In the reverse process, SD learns a denoising U-Net network $\theta$ to predict the added noise, which is expressed as:
\begin{equation}
p_{\theta}(x_{0:T}) = p_{\theta}(x_T)\prod_{i=1}^{T} p(x_{t-1} | x_t)
\end{equation}

Finally, VAE decoder $D(\cdot)$ translates the $x_0$ in latent space into pixel space as the generated image $X$

\textbf{Image Customization}
%Given a series of images depicting the same underlying concept, Image Customization aims to learn a novel concept from them. 
The key idea of Image Customization is that they inject given visual information into text tokens to generate images with a new concept. Specifically, Image Customization methods characterize and learn a new token embedding vector $V^*$ through CLIP~\cite{radford2021learning}. The training objective is expressed as:
\begin{equation}
V^*=\;\mathop{\arg\min}\limits_{V} \mathbbm{E}_{x\sim E(X), V, z\sim\mathcal{N}(0,1),t} [|\!|z-z_{\theta}(x_t, t, V)|\!|_2^2 ]
\end{equation}
After the completion of training, $V^*$ can be utilized as the novel concept embedding for the generation of customized conceptual images.

\begin{figure*}[]
\centering
\includegraphics[width=0.9\linewidth]{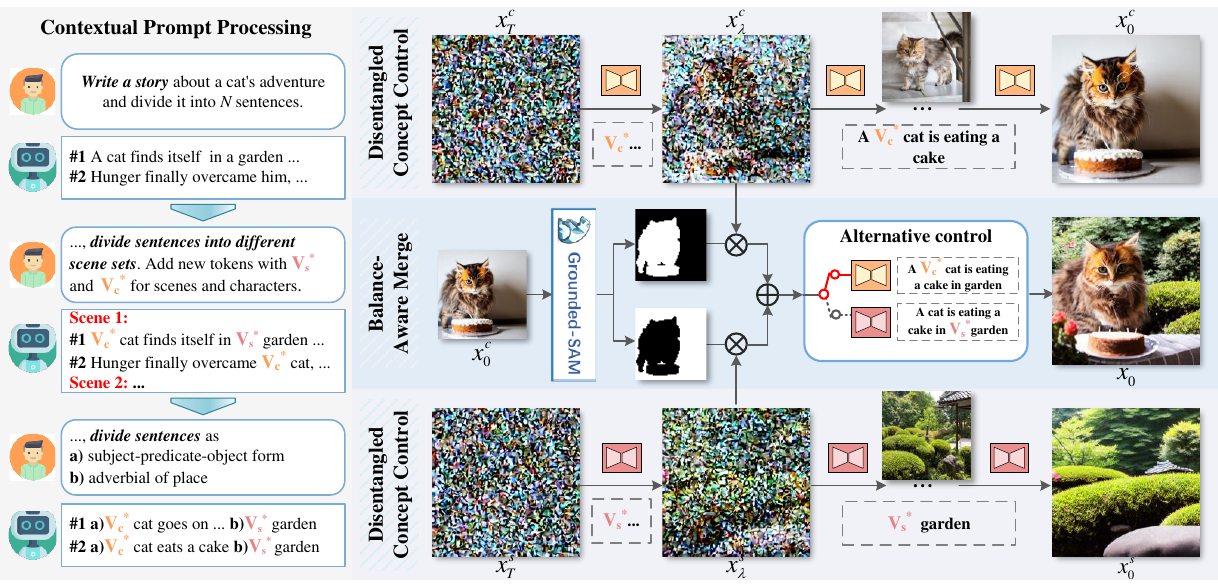}
\caption{\textbf{Overview of our Make-A-Storyboard.} \textit{Firstly}, in Contextual Prompt Processing, LLMs are employed to 
%identify specific concepts of scenes and characters that should be visually consistent.
dividing the scene information of the provided story.
\textit{Secondly}, in Disentangled Concept Control, the specific concepts of characters and scenes are independently learned for semantics preservation. \textit{Finally}, we propose Balance-Aware Merge to merge features of characters and scenes in the middle of the denoising process for the balance of visual consistency and semantics alignment. 
}
\label{pipeline}
\end{figure*}

\section{Method}

We propose a new representation form called Storyboard for Story Visualization. Storyboard involves visualizing a given story in a scene-by-scene format. In each scene, the synthesized images should maintain consistency in both scene and characters. To achieve it, we design a general framework Make-A-Storyboard as shown in Fig.~\ref{pipeline}.
We begin by dividing the scene information of the provided story by Contextual Prompt Processing. Next, within each specific scene, we decouple scene and character generation in~\ref{sec:disentangle} and then based on observation drawn from~\ref{sec:observation} we employ Balance-Aware Merge to generate harmonized images for the Storyboard in~\ref{sec:fusion}.

\subsection{Contextual Prompt Processing.}\label{sec:cpp}
As demonstrated in the left part of Fig.~\ref{pipeline}, an LLM is utilized to process story prompts considering contextual correlations among sentences.
To begin with, the LLM writes a story according to a given topic and then divides the story into $N$ discrete sentences expressed as $\mathcal{T}$.

Then we enable the LLM to identify adverbials and subjects that describe the same scenes and characters among different sentences according to contextual relationships.
For scenes, we assume that the story continues within the same scenes when the story does not explicitly change the scenes or uses pronouns referring to previously mentioned scenes.
According to the assumptions, LLMs partition sentences into scene sets based on contexts, and add a new token embedding $V_s^*$ to denote the specific scene concept for the current set.
Then in each scene, we identify the characters. And pronouns are mapped to the specific character names and introduce $V_c^*$ to represent its concept.
Through the aforementioned processing steps on sentence, we can obtain modifier story sentences as $\mathcal{T^*}$.

Finally, due to the quality degradation observed in the current T2I model when handling lengthy and literary prompts, we implement a rewriting process. Specifically, we employ the LLM to simplify the expression of each sentence, while also dividing it into two distinct components: $\mathcal{T}_s^*$ and $\mathcal{T}_c^*$. $\mathcal{T}_s^*$ is an adverbial of place, representing the previously obtained scene information. And $\mathcal{T}_c^*$ denotes the narrative content presented in a subject-verb-object form. Meanwhile, we also get the original prompt $\mathcal{T}_s$ and $\mathcal{T}_c$ without new tokens.

\subsection{Disentangled Concept Control}\label{sec:disentangle}
To keep visual consistency among characters and contextual correlation in scenes, a straightforward approach involves utilizing Image Customization methods with the capacity for fine-tuning multiple concepts, such as Custom Diffusion~\cite{kumari2023multi}. However, as shown in Fig.~\ref{Comp14}, current customization methods result in entanglement of visual features when it comes to concurrently handling both characters and scenes. Therefore, we disentangle the visual concept control of characters and scenes as shown in Fig.~\ref{pipeline}.

Specifically, given scene and character reference images, we first separately fine-tune the parameters through an Image Customization model. And we get specific U-net parameters $\theta_c$ for character and $\theta_s$ for scene as well as their modifier vector embedding $V_c^*$ and $V_s^*$. In a formal representation, character fine-tuning can be expressed as follows:
\begin{equation}
(V_c^*\!,\!\theta_c)\!=\!\mathop{\arg\min}\limits_{(V , \theta)} \mathbbm{E}_{x\sim E(X_c)\!, \!V\!,\! z\sim\mathcal{N}(0,1),t} [|\!|z-z_{\theta}(x_t, t, V)|\!|_2^2 ],
\end{equation}
where $X_c$ denotes the reference characters. The fine-tuning of scenes follows the same scheme.

\begin{figure*}[!]
\centering
\includegraphics[width=0.85\linewidth]{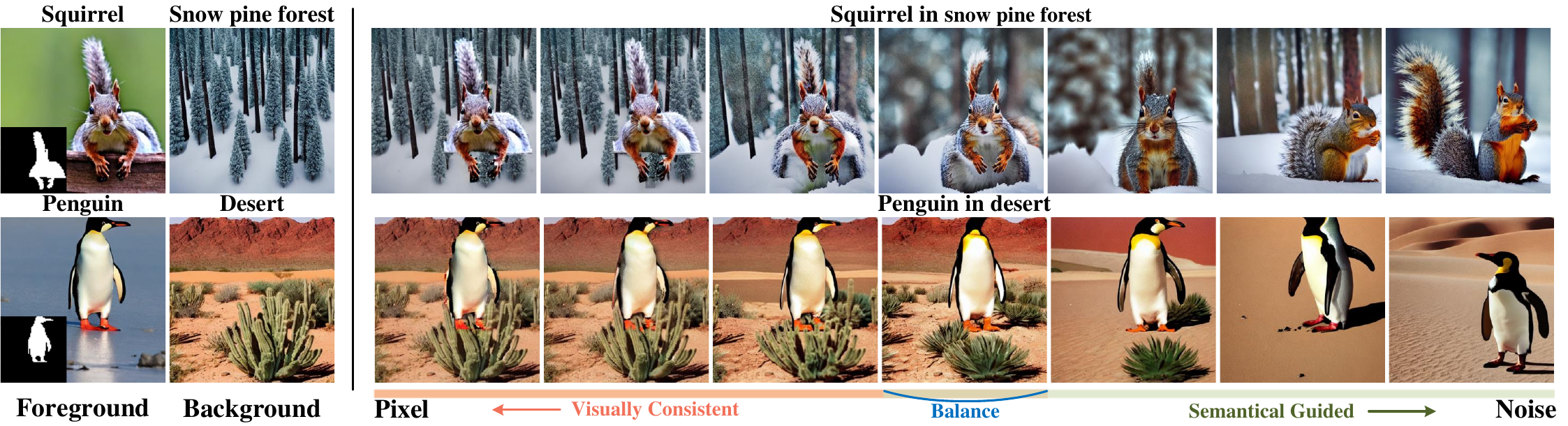}
\caption{\textbf{Observation on Feature Merge in Diffusion}. Fusing latent features of two images by spatial masks in the denoising process from pixel to noise results in a gradual transition from totally visually consistent images to totally semantically guided images.}
\label{observation}
\end{figure*}

Given $\theta_c$ equipped with the knowledge of the characters visual concepts, we could present the distributions of $x_0^c$ of characters guided by prompt $\mathcal{T}_c^*$ as:
\begin{equation}
p_{\theta_c}(x^c_{0:T}) = p_{\theta_c}(x^c_T)\prod_{i=1}^{T} p(x_{t-1}^c | x_t^c , \mathcal{T}_c^*),
\end{equation}
where $x_0^c$ represents the final latent feature 
The process for scene generation is analogous.

\subsection{Observation on Feature Merge in Diffusion}~\label{sec:observation}
Given the disentangled controlled scenes and characters, we ought to merge them into a harmonized image, while keeping their learned visual concepts.
However, simply cutting and pasting the latent code $x_0^c$ and $x_0^s$ results in the pronounced disjointed effect.
To address the issue, we conduct a crucial observation on the merge phenomena in StableDiffusion: \textit{fusing latent features of two images in the denoising process from pixel to noise results in a gradual transition from totally visually consistent images to totally semantically guided images} as shown in Fig.~\ref{observation}.

Specifically, utilizing Stable Diffusion, we generate foreground character and background scene images separately through distinct prompts for character $\mathcal{P}_c$ and scene $\mathcal{P}_s$. Then, at a certain intermediate step of the DDIM denoising process, we merge the character feature with the scene feature under the guidance of the character mask. Then we employ the composite prompt $\mathcal{P} = \mathcal{P}_c + \mathcal{P}_s$ to control the subsequent DDIM denoising steps. In this manner, the resulting images differ from those obtained by direct collage in pixel space, giving rise to a noticeable integration process.

On the one hand, when the merge process is situated closer to the pixel side, the guidance provided by the merge prompt $\mathcal{P}$ is limited. Consequently, the merged images exhibit good visual consistency respectively with the original characters and scenes.
However, near the pixel side, high-frequency information such as edges has already been largely determined, the denoising process primarily focuses on low-frequency information like textures and colors. This leads to challenges in effectively integrating scenes and characters into a harmonized one, with a rigid edge. As verified by visual results in Fig. \ref{observation}, the characters and scenes are similar to the original images but the squirrel and penguin both float abruptly in the background.

On the other hand, when the merge process is extremely close to the noise side, it can be approximated as the superposition of two Gaussian noise distributions, with the merged feature still obeying a Gaussian distribution. In such a way, the merge process approximates a new T2I synthesis guided by the merged prompt $\mathcal{P}$, resulting in minimal preservation of visual information with the original image. 
% As validated in the Fig.~\ref{observation}, the synthesized image in the noise side is a totally new image conditioned on prompts. 
Whereas, the closer the merge process is to the noise side, the less deterministic the merged feature is. Therefore, the high-frequency information and semantics of characters and scenes will be better integrated and controlled by prompts, contributing to semantical alignment with prompts.

Consequently, to trade-off between visual consistency with character and scene images and semantics alignment for reasonable images, we consider the middle stage of the denoising process as the merge stage.

\subsection{Balance-Aware Merge}~\label{sec:fusion}
Inspired by the aforementioned observations, we perform Balance-Aware Merge for scenes and characters feature to balance visual consistency and semantic guidance during the denoising process.

\begin{figure*}[]
\centering
\includegraphics[width=1\linewidth]{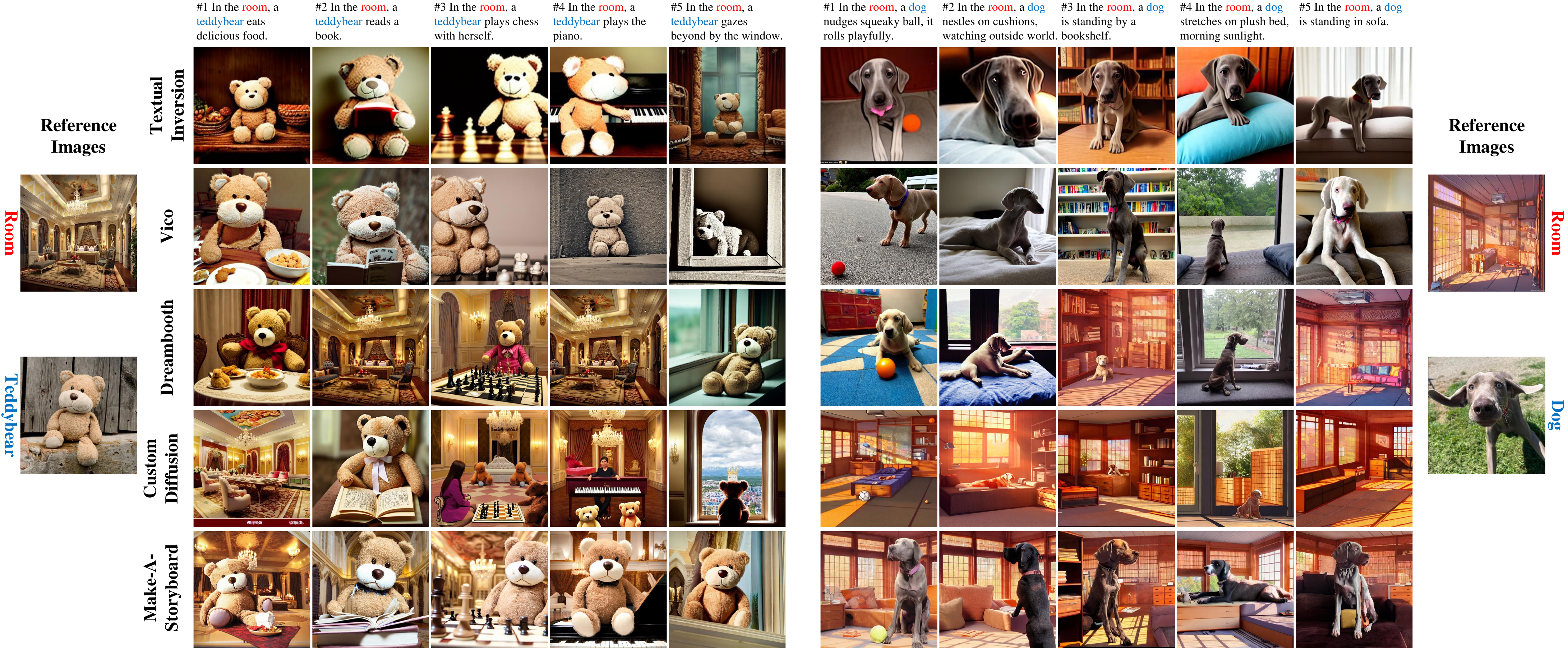}
\caption{ \textbf{Qualitative comparison with SOTA Image Customization approaches.} Comparatively, our Make-A-Storyboard stands out for ensuring textual alignment, scene consistency and character consistency.}
\label{Comp14}
\end{figure*}

In order to merge the foreground and the background, we employ a powerful segmentation model Grounded-Segment-Anything~\cite{kirillov2023segment,liu2023grounding} expressed as $SAM(\cdot)$ to obtain the mask for the foreground. To be specific, we first take the former generated character latent feature $x_0^c$ and pass it through VAE decoder $D(\cdot)$ and obtain pixel-space images $D(x_0^c)$, which is aimed at facilitating subsequent segmentation. Subsequently, we utilize $\mathcal{T}_{sam}$ to obtain the mask performed by Grounded-Segment-Anything and then resize it to fit the scale of the latent feature. The $\mathcal{T}_{sam}$ is composed of the subject and object from $\mathcal{T}_c$ which ensures that objects in the narrative story are not lost. 
% Subsequently, we resize the mask through an interpolation method to latent space as $m$, facilitating subsequent fusion. 
The process of obtaining the mask can be formalized as:
\begin{equation}
m = Resize(SAM(D(x_0^c) , \mathcal{T}_{sam}))
\end{equation}
Finally, we perform the merge in the midst of the denoising steps and get merged latent feature $x_\lambda$ according to the following equation.
\begin{equation}
x_\lambda = (1-m)\odot x_\lambda^s + m\odot x_\lambda^c,
\end{equation}
where parameter $\lambda$ represent the merging DDIM step.

After the foreground-background merge, following the approach we use in the observation, proceeding with the subsequent DDIM process only under the guidance of the unfine-tuned prompt $\mathcal{T}$ and U-net would result in a significant visual semantic shift for character and scene. Hence, in the subsequent denoising process of DDIM, it is imperative to concurrently control both the fine-tuned concepts of characters and scenes. However, extensive experiments have demonstrated that the current Image Customization methods fail to maintain high quality and decoupling when simultaneously controlling these two concepts.

\textbf{Alternate Control.} In order to prevent the confusion of visual information, we propose the Alternative Control method to completely decouple the generation of scenes and characters. Specifically, we control the denoising process by alternating between adding the trained scene embedding $V_s^*$ and the fine-tune character embedding $V_c^*$. For instance, the alternation is conducted using the following prompts ``A cat is eating a cake in $V_s^*$ garden" and ``A $V_c^*$ cat is eating a cake in garden". Furthermore, in addition to the alternating use of the corresponding embeddings, we also alternate the use of the corresponding fine-tuned U-net. The formal representation of the distribution of $x_0$ is as follows:
\begin{equation}
\begin{aligned}
p(x_{0:\lambda}) = p_{\theta}(x_\lambda)\prod_{i=2k+1,k\in N}^{\lambda-1} &p_{\theta_s}(x_{i-1} | x_i,\mathcal{T}_c+\mathcal{T}_s^*) \\
\prod_{i=2k,k\in N}^{\lambda} &p_{\theta_c}(x_{i-1} | x_i,\mathcal{T}_c^*+\mathcal{T}_s ) ,
\end{aligned}
\end{equation}
where $\mathcal{T}_c$ and $\mathcal{T}_s$ represent adverbial of place and narrative content respectively, while $\mathcal{T}_c^*$ and $\mathcal{T}_s^*$ denote the incorporation of corresponding modifier embeddings. $\theta_c$ and $\theta_s$ signify the network parameters for the character and the scene.

\section{Experiments And Analysis}

\subsection{Implementation Details}

\textbf{Datasets.}
We test our method on a total of eight concept datasets, comprising four character concepts and four scene concepts. Each concept encompasses 5-9 images. Unlike other open-domain story visualization approaches~\cite{gong2023talecrafter,liu2023intelligent}, we do not require training on large-scale datasets. If no special instructions, in quantitative comparisons, we conduct evaluations on a total of 2400 images across 5 stories, with each story comprising 20 prompts, and each prompt containing 24 samples.

\textbf{Evaluation Metrics.}
\textbf{(1)Image-alignment} We assess the alignment of the visualized story with reference images using image-to-image alignment in the CLIP image feature space~\cite{gal2022image}. We compute the image alignment with references separately for scenes and characters, following by averaging these scores to obtain the total Image-alignment \textbf{(2)Text-alignment} we evaluate the alignment of generated images with story prompts using text-to-image alignment~\cite{hessel2021clipscore} in the CLIP feature space.

\subsection{Comparison with Image Customization} \label{sec:compwithcustomized}
\textbf{Qualitative Comparisons.}
As shown in Fig.~\ref{Comp14}, we present qualitative results comparing our approach integrated into Custom Diffsion to SOTA IC methods.
Textual Inversion fails to generate high-quality images and performs poorly in both scene and character alignment. As for Vico, despite maintaining satisfactory character information, it entirely fails to simultaneously handle scene information. The problem leads to a highly fragmented narrative.
Custom Diffusion often encounters issues with character loss, and significant information loss regarding the features of both characters and scenes occurs frequently. Dreambooth is totally unable to generate a coherent composition where scenes and characters coexist harmoniously, frequently resulting in fragmented images where either scenes or characters predominate. 
Our approach addresses the entanglement issue between scene and character features, enabling the generation of stable and high-quality story visualization images. For image-alignment, our approach stands as the sole method capable of ensuring the consistency of both characters and scenes simultaneously. In terms of text-alignment, ours allows for a better focus on the details within the narrative compared with baselines, achieving improved alignment with the given prompt .

\textbf{Quantitative Comparisons.}
As revealed in Tab.~\ref{tabComp1}, ours yields the best results in both Text-alignment and Image-alignment, which is aligned with the analyses presented in the Qualitative Comparisons. Leveraging the disentangled control over scenes and characters, our method excels in adhering to prompts while preserving the visual concepts from both reference images of scenes and characters.

\textbf{User Study.}
We conduct a user study evaluation involving a total of 20 participants who provided ratings for 5 methods in 4 stories, each comprising 3 images. We utilize three evaluation metrics: Text-alignment, Image-alignment, and Overall visual storytelling Quality. For each story, participants are tasked with ranking five methods, assigning scores from 5 to 1 for each ranking. 
% A higher score indicates better performance with 5 points representing the highest rating. 
As illustrated in Tab.~\ref{tabuserstudy}, we achieve the highest scores across all three metrics, proving the effectiveness of our method.

\begin{figure}[!]
\centering
\includegraphics[width=1\linewidth]{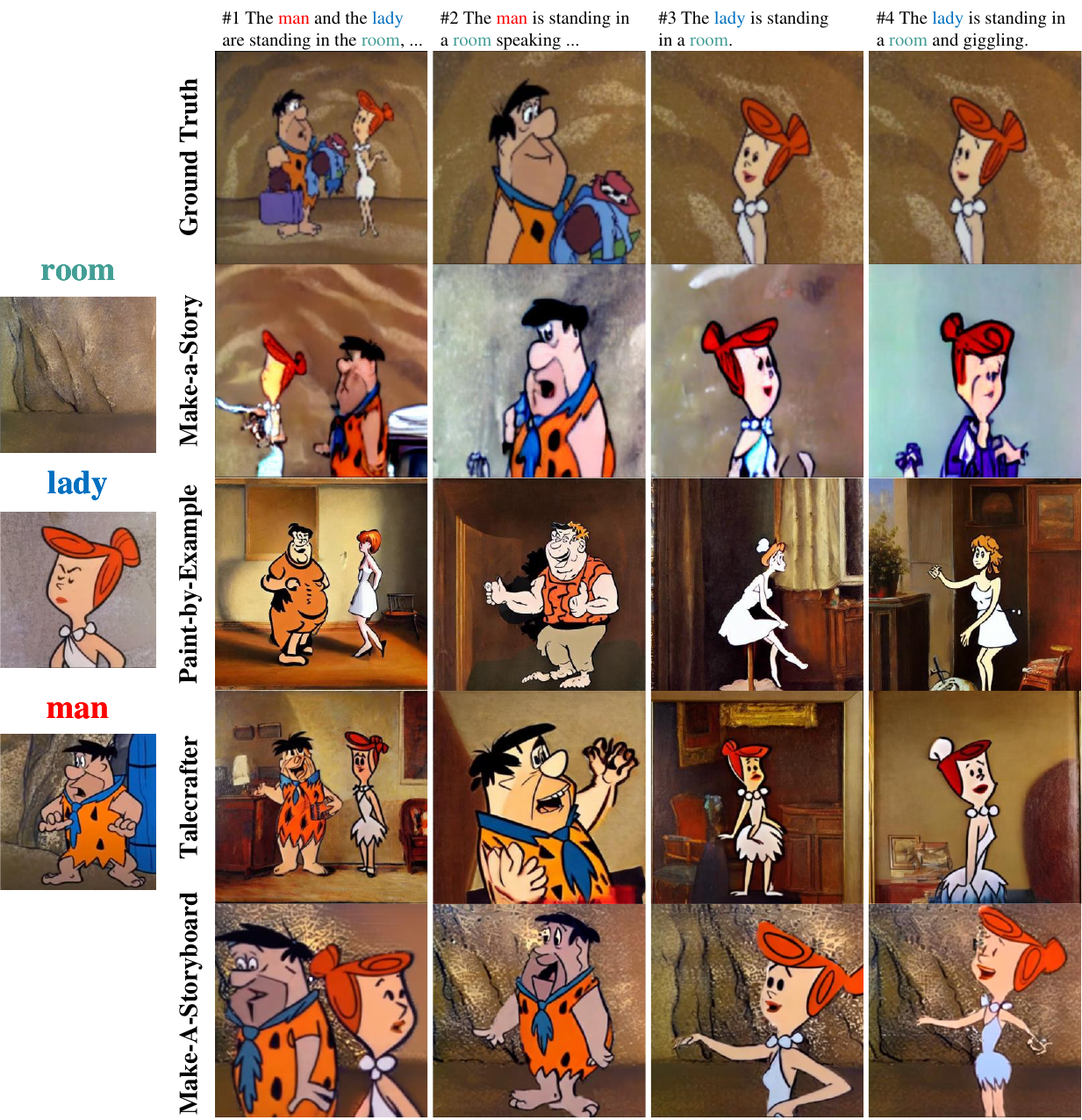}
\caption{\textbf{Qualitative comparison with SOTA story visualization.} Ours not only achieves the best visual quality but also the best alignment on both foreground and background.}
\label{trastory}
\end{figure}

\subsection{Comparison with Story Visualization Methods}
We compare our work with SOTA baselines in the field of story visualization. We use their provided display images to compare with our method for the absence of published code. For multi-character generation, we extend our framework. We disentangle multiple characters and scene generation respectively and offset character masks in the merge stage to avoid overlapping of characters.

As shown in Fig.~\ref{trastory}, it is evident that Make-a-Story and Paint-by-Example exhibit unsatisfying image quality and poor similarity. While maintaining reasonable image quality, Talecrafter entirely overlooks the scene, resulting in an oil painting style rather than cave-style in ground truth. In contrast, our approach, when be integrated into Dreambooth, demonstrates the capability to generate multiple characters while maintaining excellent consistency for both characters and scenes. It is worth noting that both Make-a-Story and Talecrafter require extensive training on large datasets, whereas our model only requires fine-tuning on a few images, significantly reducing the training cost.

\begin{table}[]
\centering
\resizebox{1\linewidth}{!}{%
\begin{tabular}{llclcl}
\hline
\multicolumn{2}{l}{Method} & \multicolumn{2}{c}{Text-alignment} & \multicolumn{2}{c}{Image-alignment} \\ \hline
\multicolumn{2}{l}{Textual Inversion} & \multicolumn{2}{c}{0.704} & \multicolumn{2}{c}{0.681} \\
\multicolumn{2}{l}{Vico} & \multicolumn{2}{c}{0.717} & \multicolumn{2}{c}{0.714} \\
\multicolumn{2}{l}{Dreambooth} & \multicolumn{2}{c}{0.735} & \multicolumn{2}{c}{0.712} \\
\multicolumn{2}{l}{Custom Diffusion} & \multicolumn{2}{c}{0.709} & \multicolumn{2}{c}{0.720} \\
\multicolumn{2}{l}{Fastcomposer} & \multicolumn{2}{c}{0.673} & \multicolumn{2}{c}{0.635} \\
\multicolumn{2}{l}{\textbf{Make-A-Storyboard}} & \multicolumn{2}{c}{\textbf{0.752}} & \multicolumn{2}{c}{\textbf{0.738}} \\ \hline
\end{tabular}%
}%
\caption{\textbf{Quantitative Comparisons with Image Customization methods on Text-alignment and Image-alignment.} Our approach outperforms SOTA methods on both metrics.}
\label{tabComp1}
\end{table}

% \usepackage{graphicx}
\begin{table}[]
\centering
\resizebox{1\linewidth}{!}{%
\begin{tabular}{llclcc}
\hline
\multicolumn{2}{l}{Method} & \multicolumn{2}{c}{Text-alignment} & Image-alignment & Overall quality \\ \hline
\multicolumn{2}{l}{Textual Inversion} & \multicolumn{2}{c}{1.96} & 1.98 & 1.66 \\
\multicolumn{2}{l}{Vico} & \multicolumn{2}{c}{3.33} & 2.73 & 3.13 \\
\multicolumn{2}{l}{Dreambooth} & \multicolumn{2}{c}{3.29} & 2.86 & 2.98 \\
\multicolumn{2}{l}{Custom Diffusion} & \multicolumn{2}{c}{2.10} & 2.84 & 2.71 \\
\multicolumn{2}{l}{\textbf{Make-A-Storyboard}} & \multicolumn{2}{c}{\textbf{4.31}} & \textbf{4.60} & \textbf{4.51} \\ \hline
\end{tabular}%
}
\caption{\textbf{User study for comparison with Image Customization methods on Text-Image alignment, Image-Image alignment, and Overall quality.} A Higher score indicates better performance, with 5 points representing the highest rating.
}
\label{tabuserstudy}
\end{table}

\begin{table}[]
\centering
\resizebox{1\linewidth}{!}{%
\begin{tabular}{llclclclcl}
\hline
\multicolumn{2}{l}{Method} & \multicolumn{2}{c}{Text-alignment} & \multicolumn{2}{c}{Image-alignment} \\ \hline
\multicolumn{2}{l}{Textual Inversion} & \multicolumn{2}{c}{0.704} & \multicolumn{2}{c}{0.681}  \\
\multicolumn{2}{l}{\textbf{Textual Inversion + Ours}} & \multicolumn{2}{c}{\textbf{0.738}} & \multicolumn{2}{c}{\textbf{0.692}} \\ \hline
\multicolumn{2}{l}{Vico} & \multicolumn{2}{c}{0.717} & \multicolumn{2}{c}{0.714}  \\
\multicolumn{2}{l}{\textbf{Vico + Ours}} & \multicolumn{2}{c}{\textbf{0.718}} & \multicolumn{2}{c}{\textbf{0.724}}  \\ \hline
\multicolumn{2}{l}{Dreambooth} & \multicolumn{2}{c}{0.735} & \multicolumn{2}{c}{0.712}  \\
\multicolumn{2}{l}{\textbf{Dreambooth + Ours}} & \multicolumn{2}{c}{\textbf{0.739}} & \multicolumn{2}{c}{\textbf{0.731}} \\ \hline
\multicolumn{2}{l}{Custom Diffusion} & \multicolumn{2}{c}{0.709} & \multicolumn{2}{c}{0.720} \\
\multicolumn{2}{l}{\textbf{Custom Diffusion + Ours}} & \multicolumn{2}{c}{\textbf{0.752}} & \multicolumn{2}{c}{\textbf{0.738}} \\ \hline
\end{tabular}%
}
\caption{  \textbf{Quantitative results on generalization of our Make-A-Storyboard.} Our general framework can be combined with current Image Customization methods, achieving improvements in both Text-alignment and Image-alignment metrics.}
\label{tabplp}
\end{table}

\begin{figure*}[]
\centering
\includegraphics[width=0.945\linewidth]{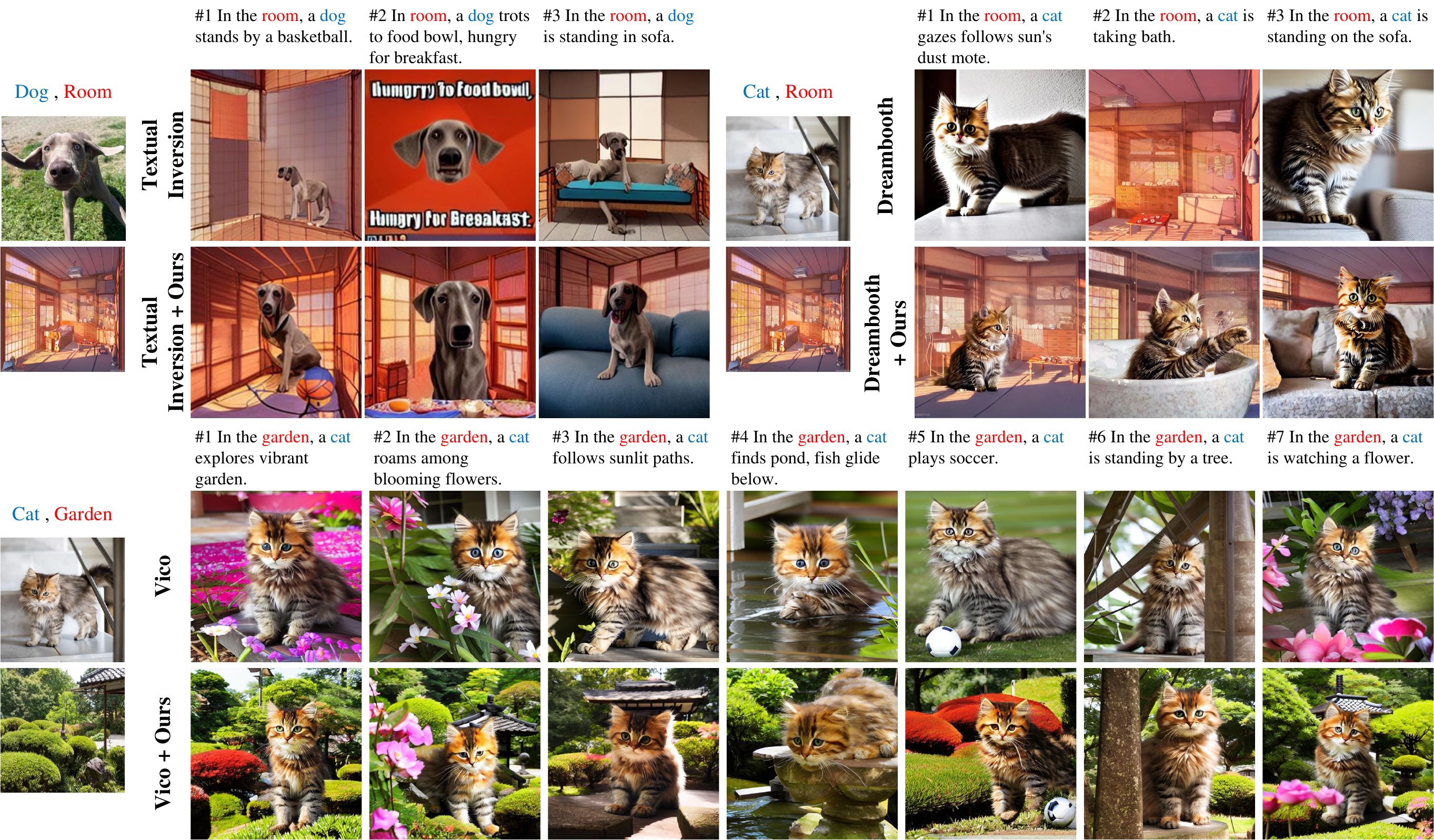}
\caption{\textbf{Qualitative Comparison on generalization of our Make-A-Storyboard.} Our method can be seamlessly integrated into Image Customization methods, which significantly enhances their capability in simultaneously controlling scene and character consistency.}
\label{compplp}
\end{figure*}

\begin{figure}[]
\centering
\includegraphics[width=0.91\linewidth]{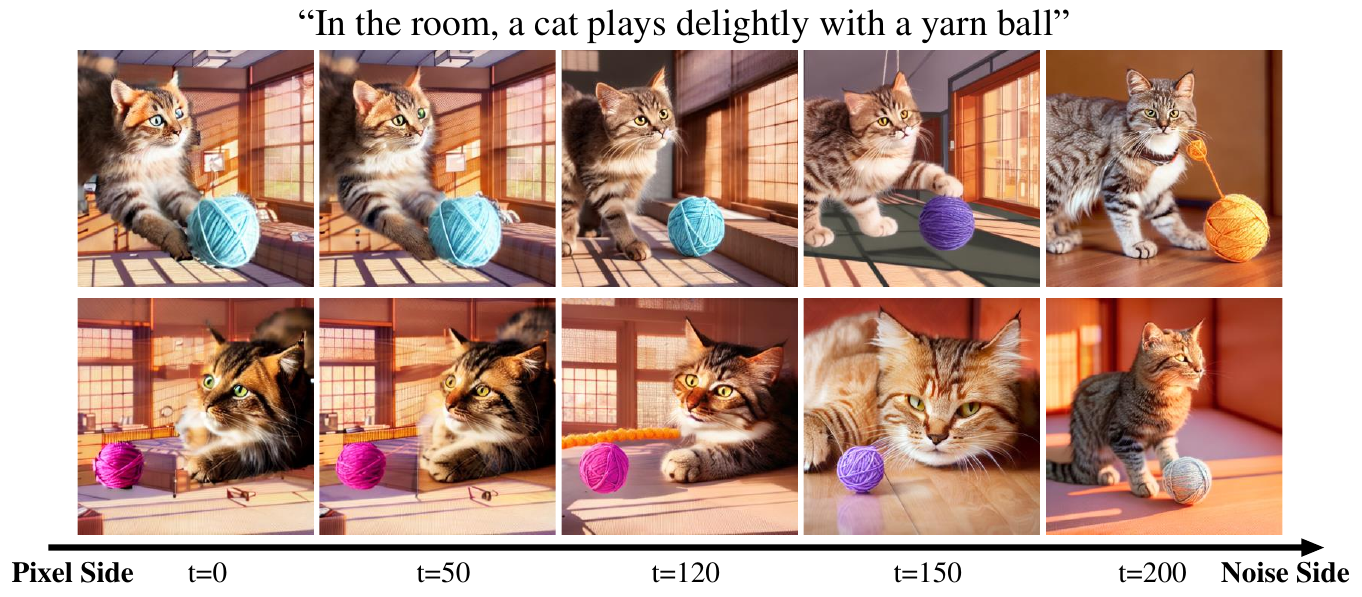}
\caption{ \textbf{Effect of different merge timesteps $\gamma$.} At the timestep of 120, the merged results trade-off between visual consistency and image harmonization.}
\label{ablation1}
\end{figure}

\subsection{Generalization on Image Customization}
% Since our approach emphasizes the decoupling of character and scene, our framework can be seamlessly integrated into Image Customization methods. 
To assess the generalization capability of our method, we integrate our method with four mainstream Image Customization methods, Textual Inversion, Vico, Dreambooth, and Custom Diffusion. The qualitative and quantitative comparisons are respectively shown in Fig.~\ref{compplp} and Tab.~\ref{tabplp}.

As shown in Tab.~\ref{tabplp}, by leveraging our framework, all methods have demonstrated improvements in both Text-alignment and Image-alignment, which proves the generalization and effectiveness of our framework. Notably, applying our framework respectively to Textual Inversion and CustomDiffusion gives substantial increase by $0.034$ and $0.043$ in Text-alignment.

Visual results shown in Fig.~\ref{compplp} further verify the generalization of ours.
Specifically, applying ours on Textual Inversion effectively addresses its issue of low generation quality. Meanwhile, Vico with our framework significantly enhances background consistency while maintaining basic character consistency. 
For DreamBooth, integrating our framework resolves the issue of unstable customization generation results when confronted with multiple concepts of scenes and characters.

\begin{figure}[]
\centering
\includegraphics[width=0.91\linewidth]{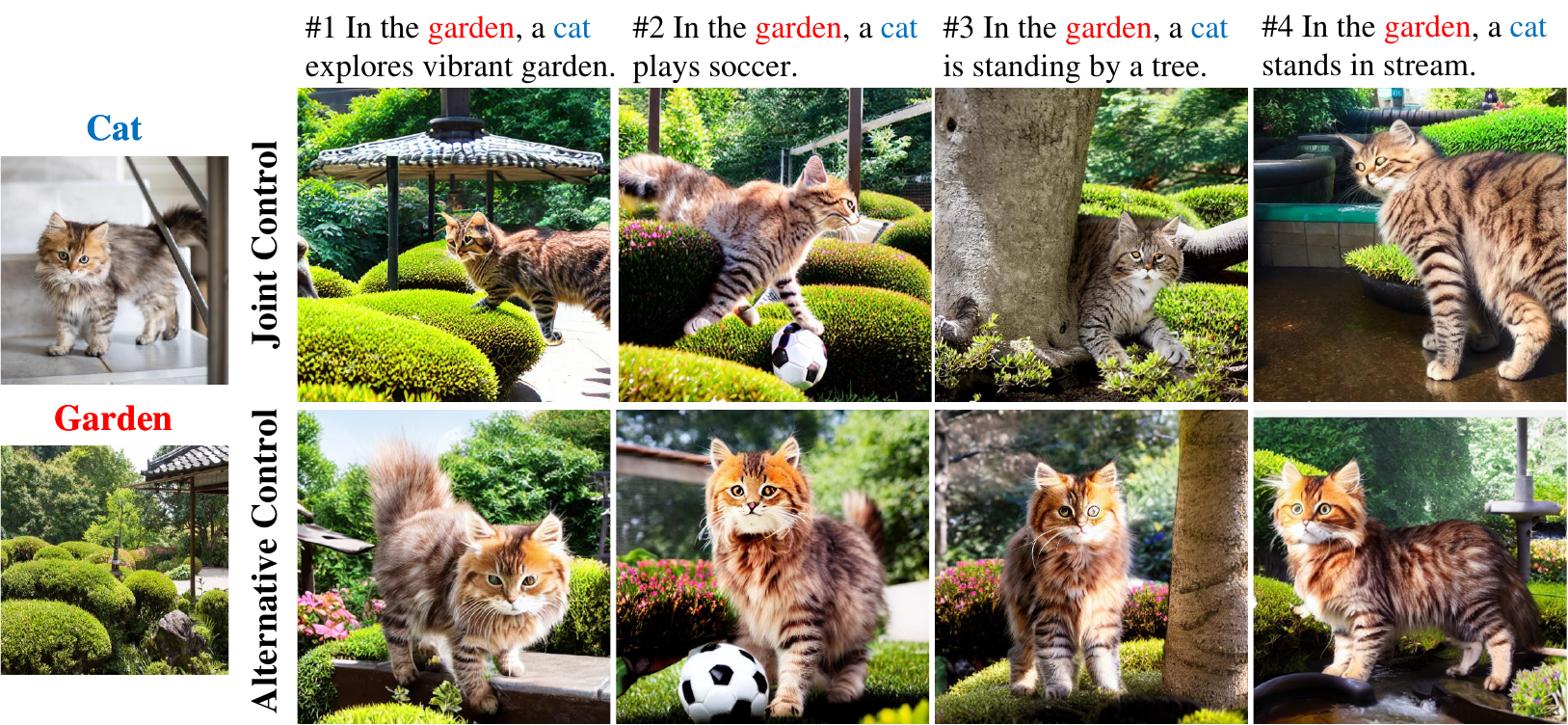}
\caption{\textbf{Effect of Alternative Control.} Compared with Joint Control of ~\cite{kumari2023multi}, Alternative Control ensures the visual consistency of both characters and scenes .}
\label{alternative_control}
\end{figure}

\subsection{Ablation Study and Analysis}
\textbf{Effect of different merge timesteps.} We study the effect of different merge timesteps $\gamma$ with DDIM sampling consisting of 200 steps. As shown in Fig.~\ref{ablation1}, 
setting the merge DDIM sampling step $\gamma$ to approximately 120 yields satisfying results, achieving an equilibrium between image coherence and alignment with reference images.

\textbf{Effect of Alternative Control.} As illustrated in Fig.~\ref{alternative_control}, we ablate the Alternative Control to show its effectiveness. We compare our approach with Joint Control which involves simultaneous fine-tuning of both scene and character concepts. %Scene Control and Character Control lead to semantic shifts in another concept while 
Joint Control results in a decrease in image quality and alignment. Comparatively, Alternative Control enables the simultaneous preservation of scene and character.

\section{Conclusion}
In this paper, we present a coherent representation form for Story Visualization called Storyboard, which unfolds story contents scene by scene. For Storyboard, we propose a general framework Make-A-Storyboard, which separately controls visual consistency of characters and scenes and merges them into harmonized images. Ours not only surpasses the SOTA in Story Visualization but also could be seamlessly integrated into Image Customization methods. 

%%%%%%%%% REFERENCES
\newpage
{\small
\bibliographystyle{ieee_fullname}
\bibliography{egbib}
}

\newpage
\begin{appendices}

In Section~\ref{sec:A}, we present additional experimental results showcasing our approach. Section~\ref{sec:B} provides implementation details aimed at improving the reproducibility of our method. Subsequently, in Section~\ref{sec:C}, we discuss limitations of our approach. Finally, in Section~\ref{sec:D}, we delve into the application and societal impact of our method.

% \tableofcontents

\section{Additional Experiment Results}
\label{sec:A}

\subsection{Visual Results of Storyboarder}
\begin{figure*}[]
\centering
\includegraphics[width=1\linewidth]{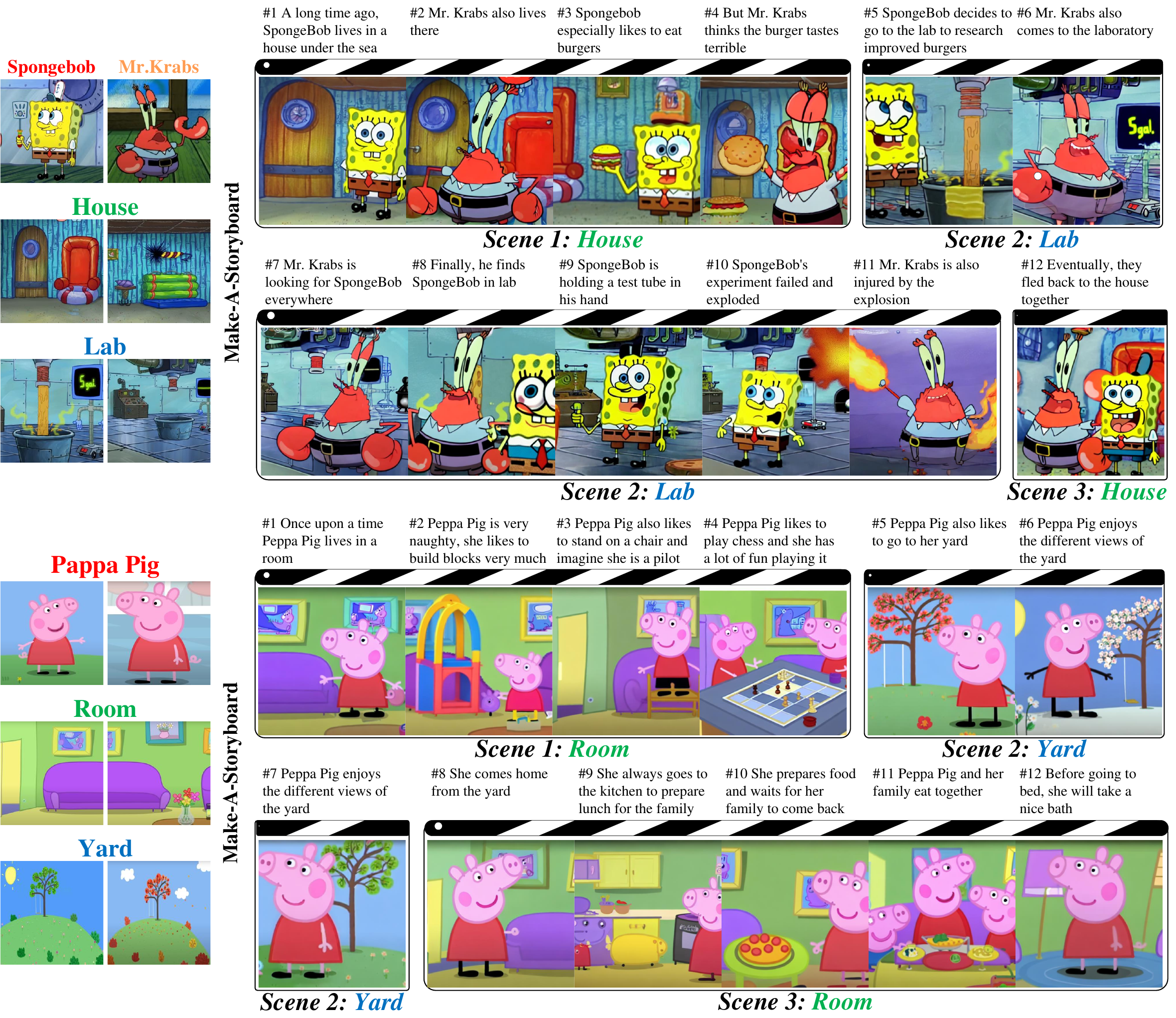}
\caption{\textbf{More visual results for Storyboard.}  The storyboard proposed by us can significantly enhance the narrative of Story Visualization.
}
\label{extra_show}
\end{figure*}

To showcase the performance of our method on longer and more complex stories, we employ our method, showcasing two examples of storyboards. As shown in Fig.~\ref{extra_show}, our method adapts to a wide range of complex scenes and characters. It ensures consistency in the visual depiction as well as in the alignment between visual and textual prompt. Additionally, by narrating the storyboard scene by scene, it greatly enhances the narrative of story visualization.

\subsection{Multi-Character Story Visualization}
Thanks to our framework's utilization of decoupling control, it can accommodate multi-concept Image Customization methods, thereby achieving a broader spectrum of story visualization. As illustrated in the Fig.~\ref{image-compo}, we demonstrate the results of our method in multi-concept story visualization when employing Custom Diffusion. It is evident that our approach maintains a commendable consistency between the generated images and the reference images even when incorporating multiple concepts. 

\subsection{Quantitative Analysis}

\begin{figure}[]
\centering
\includegraphics[width=1\linewidth]{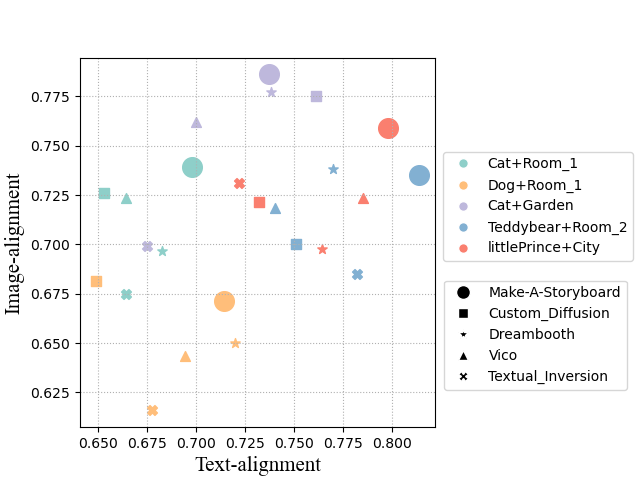}
\caption{\textbf{Quantitative results of different stories.} Distinct colors symbolize various stories, and our method (depicted by circles) consistently positions itself in the upper-right quadrant across diverse stories in comparison to other baselines. This trend suggests that our approach achieves superior outcomes concerning both Image-alignment and Text-alignment. 
}
\label{new_satter}
\end{figure}

We compare our method with SOTA using 2400 images across 5 main themes. As depicted in the Fig.~\ref{new_satter}, we display numerical comparisons for each individual theme concerning Text-alignment and Image-alignment. It is evident that for each distinct color-coded story, our method consistently occupies the top-right position relative to other approaches, demonstrating its superiority over the baselines in both evaluation metrics.

\subsection{Attention Visualization Analysis}

\begin{figure}[]
\centering
\includegraphics[width=1\linewidth]{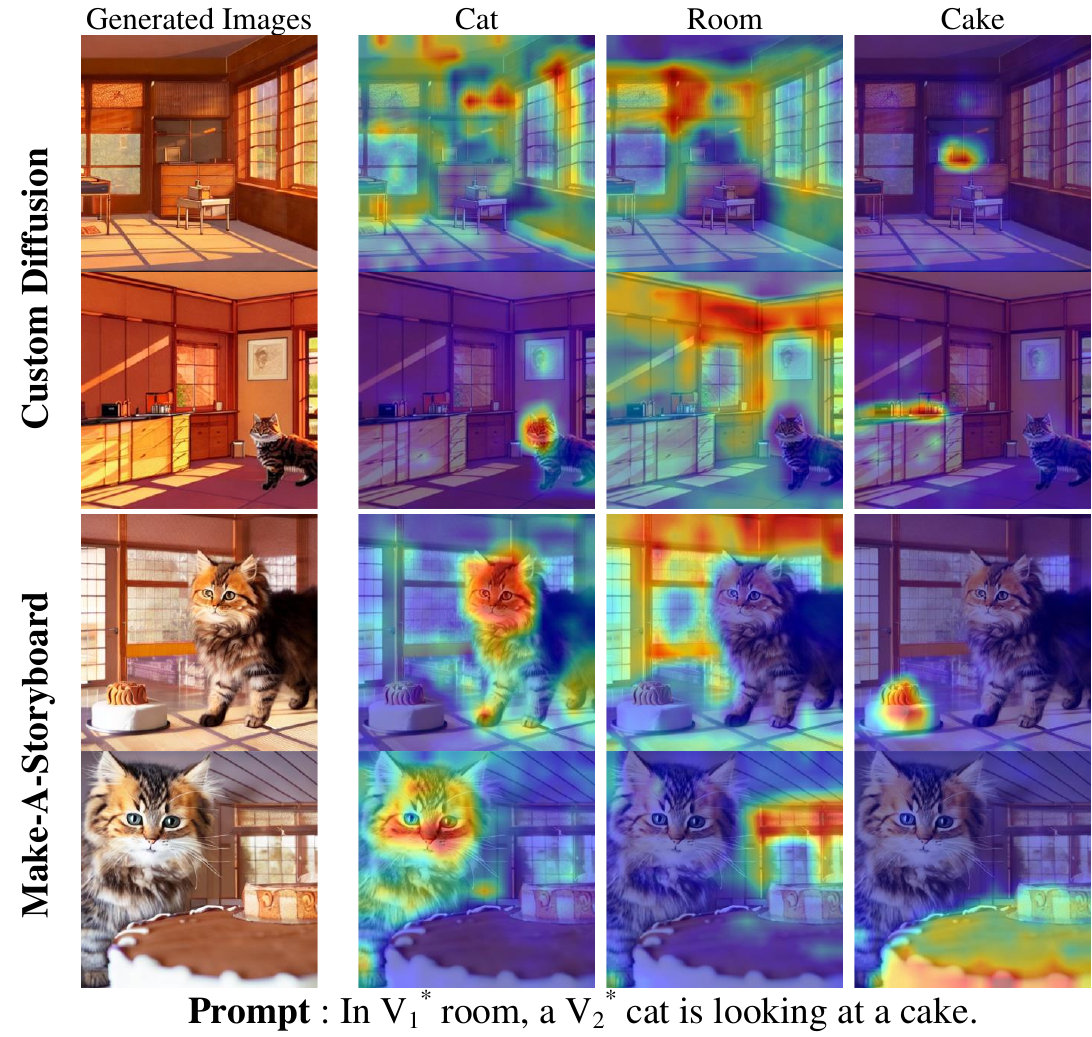}
\caption{\textbf{Attention map visualization of Custom Diffusion and Make-A-Storyboard.} Custom Diffusion encounters instances of cross-attention confusion and visual concept coupling when simultaneously fine-tuning scenes and characters. And our approach addresses these issues through a decoupling method.
}
\label{app_attention}
\end{figure}

To further illustrate the effectiveness of our method, we visualize the cross-attention maps of Custom Diffusion and Make-A-Storyboard during the generation process, as depicted in the Fig.~\ref{app_attention}.  
As shown in the first row, due to the particularity of scenes as a global concept, Custom Diffusion leads to confusion in the cross-attention between characters and scenes, resulting in frequent missing characters. Furthermore, as demonstrated in the second row, even when Custom Diffusion accurately localizes cross-attention, its ability to preserve the characters concerning the reference image is poor. This is due to the coupling between characters and scenes during the fine-tuning process, resulting in the loss of visual information.

Moreover, as depicted in the last column of the figure, scene concepts further affects Custom Diffusion's ability to maintain the prompt. In contrast, our method addresses the issues of character missing and coupling between character and scene concepts through a decoupling control and then merge approach, achieving improved image-alignment and text-alignment.

\subsection{Comparison with Image Composition}

\begin{figure*}[]
\centering
\includegraphics[width=1\linewidth]{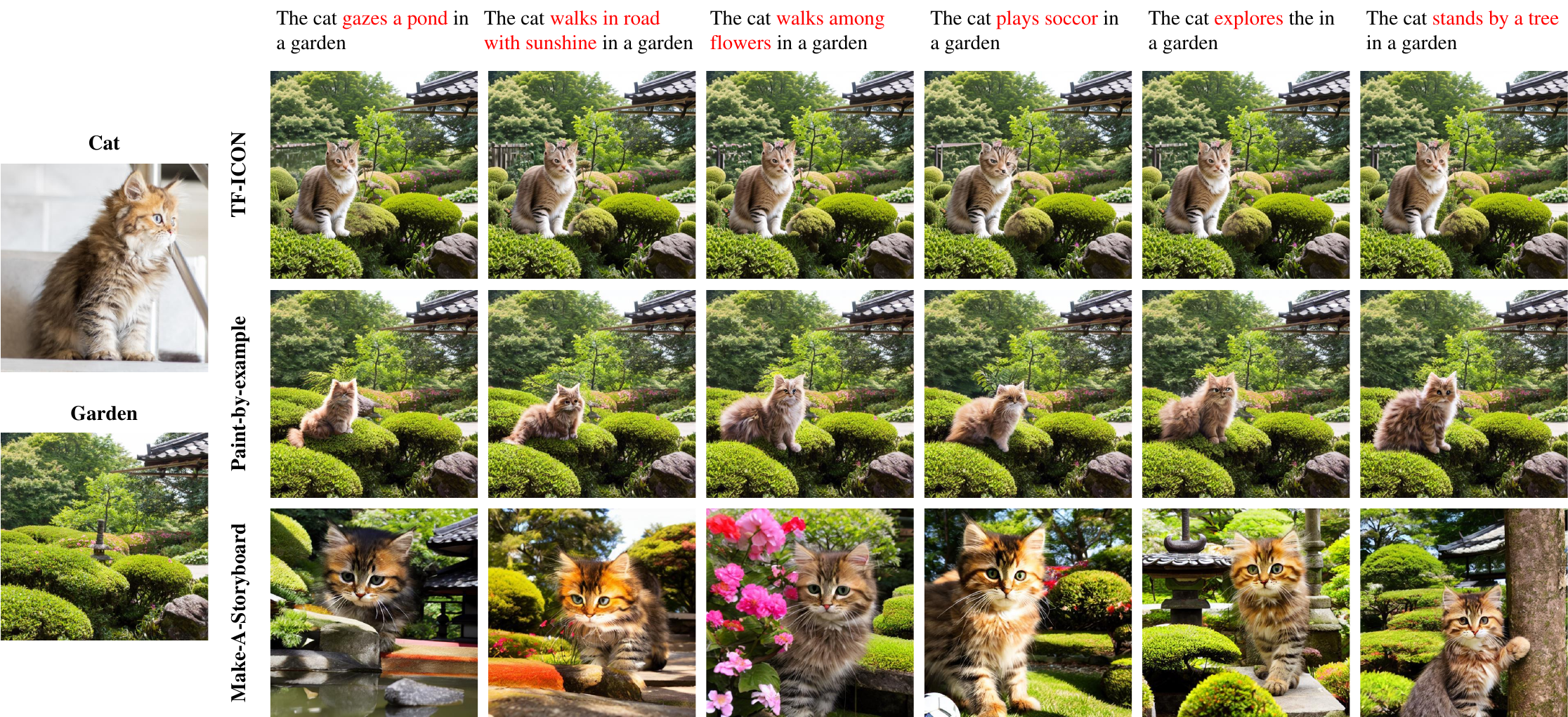}
\caption{\textbf{Comparison with SOTA Image Composition methods.} Both TF-ICON and Paint-by-example demonstrate an inability to effectively control prompts and entirely alter the actions of the reference. Our method achieves superior Text-alignment and generates a diverse range of images.
}
\label{image-compo}
\end{figure*}

Due to the similarity between our method and the concept of Image Composition, we conduct a comparison with the current SOTA image composition methods, as depicted in the Fig.~\ref{image-compo}.
Regarding Text-alignment, while TF-ICON~\cite{lu2023tf} allows specifying a prompt during the image composition process, its control over the prompt is limited, thereby almost entirely failing to maintain consistency between character actions and the prompt. Conversely, Paint-by-example~\cite{yang2023paint} lacks the ability to specify a prompt, making it unsuitable for meeting the requirements of story generation tasks. In contrast, our method, compared to image composition methods, can accommodate complex prompts and achieve superior Text-alignment.
Concerning Image-alignment, it is observable that both TF-ICON and Paint-by-example methods struggle to maintain consistency in characters and result in a significant decrease in the generated cat's quality. Meanwhile, our method not only maintains excellent consistency between characters and scenes but also offers higher flexibility and diversity.

\subsection{Comparison with Image Customization}
\begin{figure*}[]
\centering
\includegraphics[width=1\linewidth]{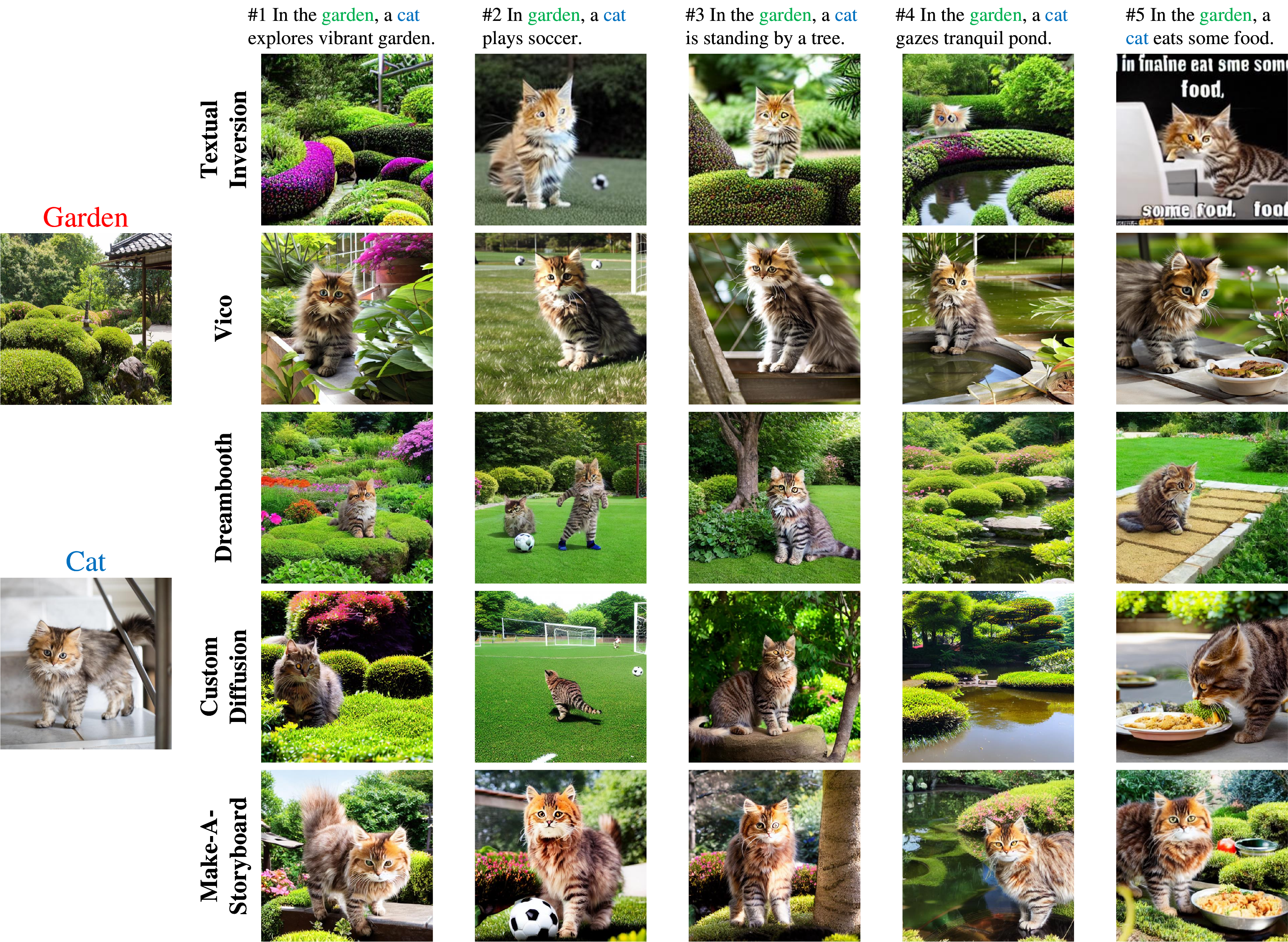}
\caption{\textbf{More comparison with SOTA Image Customization methods.} Our method significantly outperforms other Image Customization methods in both Image-alignment and Text-alignment.
}
\label{extra_comp_1}
\end{figure*}

\begin{figure*}[]
\centering
\includegraphics[width=1\linewidth]{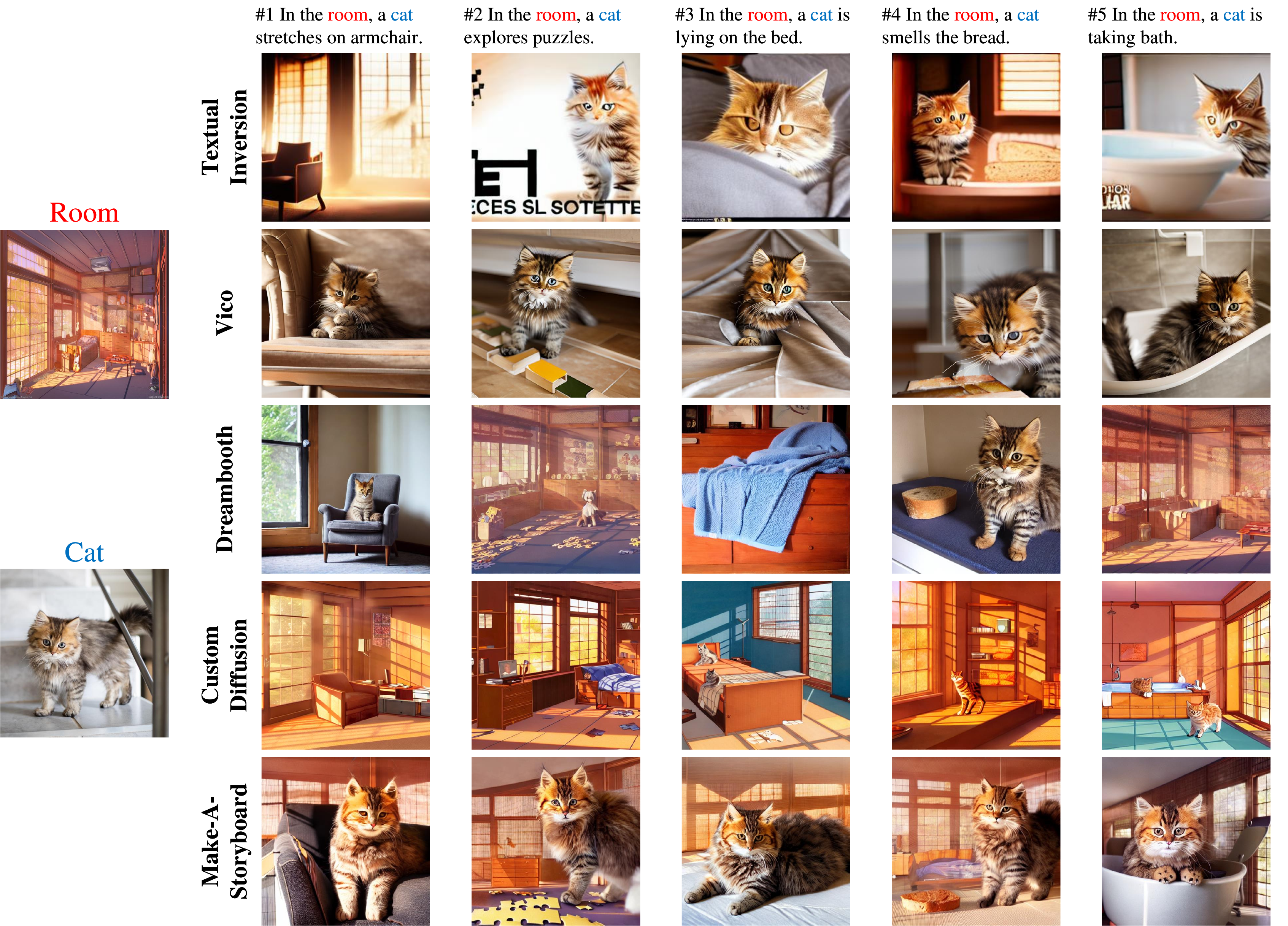}
\caption{\textbf{More comparison with SOTA Image Customization methods.} Our method significantly outperforms other Image Customization methods in both Image-alignment and Text-alignment.
}
\label{extra_comp_2}
\end{figure*}

Under the concept of storyboards, we present a more comprehensive comparison with SOTA Image Customization methods, as illustrated in the Fig.~\ref{extra_comp_1} and Fig.~\ref{extra_comp_2}.

\section{Details}
\label{sec:B}

\textbf{Different merge steps.}
When applying our framework to different image customization methods, we employ slightly varied merge steps $\lambda$ to balance image coherence and alignment. Specifically, through empirical experimentation, our framework utilizes $\lambda=120$ for Custom Diffusion and Textual Inversion, $\lambda=100$ for Dreambooth, and $\lambda=140$ for Vico.

\textbf{Comparison method implementation details.}
 We compare our method with SOTA Image Customization methods, including methods capable of multi-object fine-tuning, such as Custom Diffusion~\cite{kumari2023multi} and Fastcomposer~\cite{xiao2023fastcomposer}, as well as methods for single-object fine-tuning, such as Dreambooth~\cite{ruiz2023dreambooth}, Vico~\cite{hao2023vico}, and Textual Inversion~\cite{gal2022image}.
To concurrently fine-tune the models for both scenes and characters, we implement certain modifications in their training approach. Specifically, for Textual Inversion, we conduct separate training sessions for individual concepts, subsequently integrating them. In the case of Dreambooth, we amalgamate the training datasets and concurrently trained two concepts. Regarding Custom Diffusion, we adhere to the Custom Diffusion optimization training method to mitigate co-training bad effects. Lastly, with regard to Vico, our training only focuses on the character concept.Besides, all baselines adhere to the hyperparameters and regularization datasets as claimed in their papers.

\textbf{Multi-character fine-tuning implementation details.}
We use two distinct multi-character story visualization methods. In Section~\ref{Mul-C}, we initially merge the guitar and dog concept using Custom Diffusion's optimization approach, followed by leveraging our framework to achieve multi-character generation.
For multi-character story visualization on single-concept Image Customization method like Dreambooth, we conduct fine-tuning and generation for each character separately. During the merging phase, we translate the images and masks of multiple characters in opposite directions to prevent potential character disappearance caused by extensive overlap.
Post-merging, we utilize alternative control for image alignment in three concepts: Character 1, Character 2, and the Scene. This control manages the alignment of the generated images within these conceptual categories.

\textbf{Other Details.}
All images generated by other models are of size $512 \times 512$ except for the images generated by the Textual Inversion, which has a size of $256 \times 256$. Besides, all experiments are conducted on NVIDIA RTX 3090 GPU.

\section{Limitations}
\label{sec:C}

Although our method of decoupling generation and re-fusion ensures the harmonization of characters with scenes, there are occasions when the excessive volume of characters makes it challenging to maintain a reasonable layout in the generated images. 
Future improvements might involve incorporating size and positional control mechanisms during the merge process.

\section{Application and Social Impact}
\label{sec:D}
Regarding the scalability of our approach, benefiting from the decoupled control of our methodology, advancement in Image Customization will further extend the scope of our method. Moreover, our insightful observation on Stable Diffusion exhibits broad applicability. This can be further harnessed for tasks such as multi-concept Image Customization and Image Composition, achieving a finer equilibrium between visual consistency and image harmonization.
In terms of the societal impact of our methodology,
the new representative form of story visualization called Storyboard could better illustrate the development and narrative of a story, thus delivering a more immersive and intuitive reading experience for readers.
Therefore, we firmly believe that our methodology will propel the further evolution of illustrated storybooks.

\label{Mul-C}
\begin{figure*}[]
\centering
\includegraphics[width=1\linewidth]{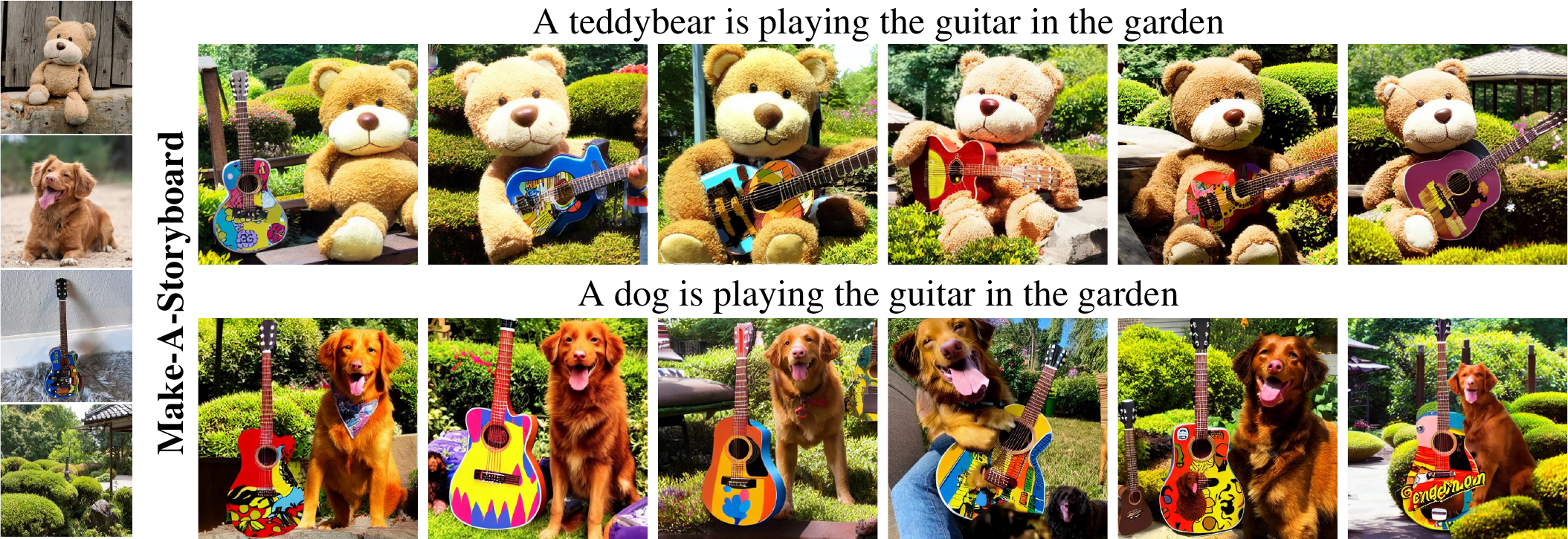}
\caption{\textbf{Multi-Character Story Visualization results.} Our approach can also simultaneous generate two characters (objects) with a specific style and a scene, encompassing three concepts for visual storytelling.
}
\label{multi_concept}
\end{figure*}

% This inspires us to consider incorporating layout control methods in the future to expand the scope of visualization across multiple concepts.

\end{appendices}
\end{document}